\documentclass[conference]{IEEEtran}
\IEEEoverridecommandlockouts
\usepackage{cite}
\usepackage{listings}
\usepackage{amsmath,amssymb,amsfonts}
\usepackage{algorithmic}
\usepackage{graphicx}
\usepackage{textcomp}
\usepackage{xcolor}
\def\BibTeX{{\rm B\kern-.05em{\sc i\kern-.025em b}\kern-.08em
    T\kern-.1667em\lower.7ex\hbox{E}\kern-.125emX}}
\begin{document}

\title{Comparative Analysis of Linear Regression, Gaussian Elimination, and LU Decomposition for CT Real Estate Purchase Decisions\\
}

\author{\IEEEauthorblockN{1\textsuperscript{st} Xilin Cheng}
\IEEEauthorblockA{\textit{Harrisburg University of Science and Technology} \\
\textit{Department of Computer Science}\\
Harrisburg, PA, USA \\}
}

\maketitle

\begin{abstract}
This paper presents a comprehensive evaluation of three distinct computational algorithms applied to the decision-making process of real estate purchases. Specifically, we analyze the efficacy of Linear Regression from Scikit-learn library, Gaussian Elimination with partial pivoting, and LU Decomposition in predicting the advisability of buying a house in the State of Connecticut based on a set of financial and market-related parameters. The algorithms' performances were compared using a dataset encompassing town-specific details, yearly data, interest rates, and median sale ratios. Our results demonstrate significant differences in predictive accuracy, with Linear Regression and LU Decomposition providing the most reliable recommendations and Gaussian Elimination showing limitations in stability and performance. The study's findings emphasize the importance of algorithm selection in predictive analytic and offer insights into the practical applications of computational methods in real estate investment strategies. By evaluating model efficacy through metrics such as R-squared scores and Mean Squared Error, we provide a nuanced understanding of each method's strengths and weaknesses, contributing valuable knowledge to the fields of real estate analysis and predictive modeling.
\end{abstract}


\section{Introduction}
The decision to invest in real estate hinges on accurate and timely analysis of market trends, interest rates, and property valuation. As such, predictive models that can process various financial indicators and suggest optimal buying opportunities are of significant interest to investors and analysts alike. This research is motivated by the need for reliable computational models that can assist in making informed real estate purchase decisions.

In recent years, the application of machine learning and numerical methods in financial decision-making has seen notable advancement. Linear Regression (LR) is a well-established statistical method for predicting a dependent variable based on independent variable(s), and it has been extensively applied within the real estate domain. However, traditional numerical methods, such as Gaussian Elimination (GE) with partial pivoting and LU Decomposition (LU), have been less explored in this context, despite their potential in solving linear systems.

This study seeks to fill the gap by implementing and comparing the performance of LR, GE, and LU in the specific task of predicting real estate purchase recommendations. We define our problem as the binary classification task of determining whether to buy a property, encapsulated by a 'Buy Recommendation' label, which is influenced by a set of predictors including town demographics, historical interest rates, and median sale ratios.

Our approach involves constructing a dataset from historical real estate transactions and financial rates, which is then processed through each of the three algorithms. We employ rigorous evaluation metrics, such as R-squared scores and Mean Squared Error, to assess the accuracy and reliability of each method's predictions.

The contribution of this paper is twofold: first, it provides a direct performance comparison of three distinct algorithms in the context of real estate decision-making, and second, it offers insights into the practical implications of selecting appropriate computational methods for predictive analytics in the housing market. We anticipate that our findings will be of particular interest to both practitioners in the real estate industry and researchers in the field of financial modeling. 

\section{Data Preprocessing}

\begin{figure}[htbp]
\centerline{\includegraphics[width=0.5\linewidth]{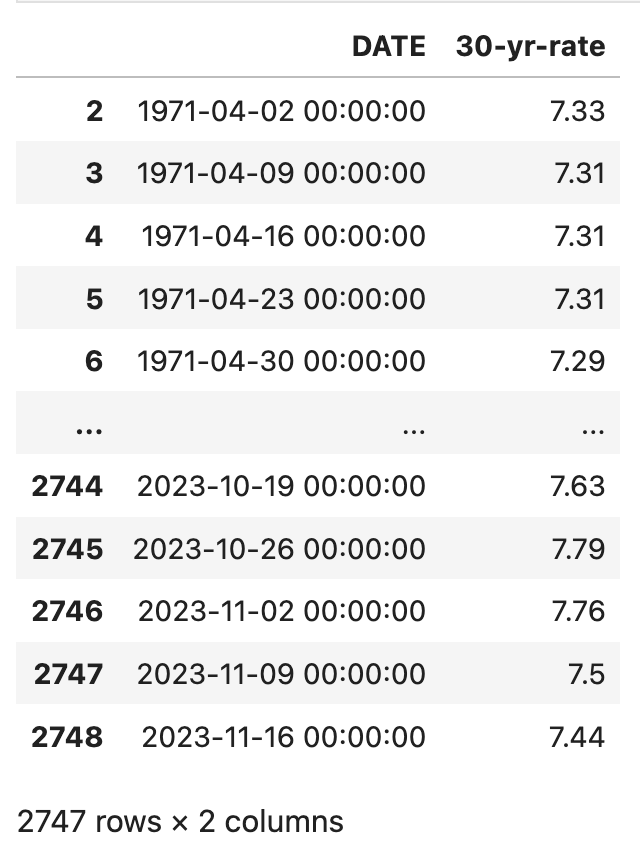}}
\caption{Raw National Mortgage Interest Rate}
\label{fig1}
\end{figure}

\begin{figure}[htbp]
\centerline{\includegraphics[width=0.5\linewidth]{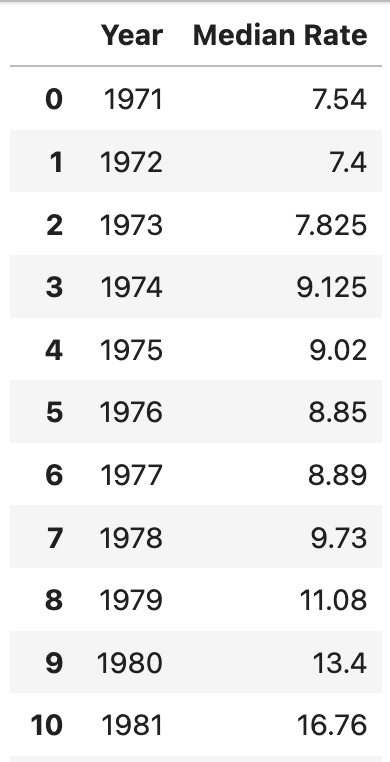}}
\caption{Median Mortgage Rate By Year}
\label{fig2}
\end{figure}

The data preprocessing stage is crucial in ensuring that the data used in the analysis is clean, relevant, and structured appropriately for the computational models. For this study, the dataset encompassing real estate sales in Connecticut from 2001 to 2020 was sourced from the Data.gov catalog.\cite{b1} Additionally, national mortgage interest rates were obtained from Freddie Mac's Primary Mortgage Market Survey®.\cite{b2} The following preprocessing steps were undertaken to prepare the dataset for the comparative analysis of predictive modeling algorithms:

1. Interest Rate Aggregation: The national mortgage interest rate data (Figure~\ref{fig1}) were aggregated to calculate the median rate for each year. This step involved a group-by operation on the 'Year' column of the interest rate dataset and applying the median function to the 'Interest Rate' column. (Figure~\ref{fig2})

\begin{figure}[htbp]
\centerline{\includegraphics[width=\linewidth]{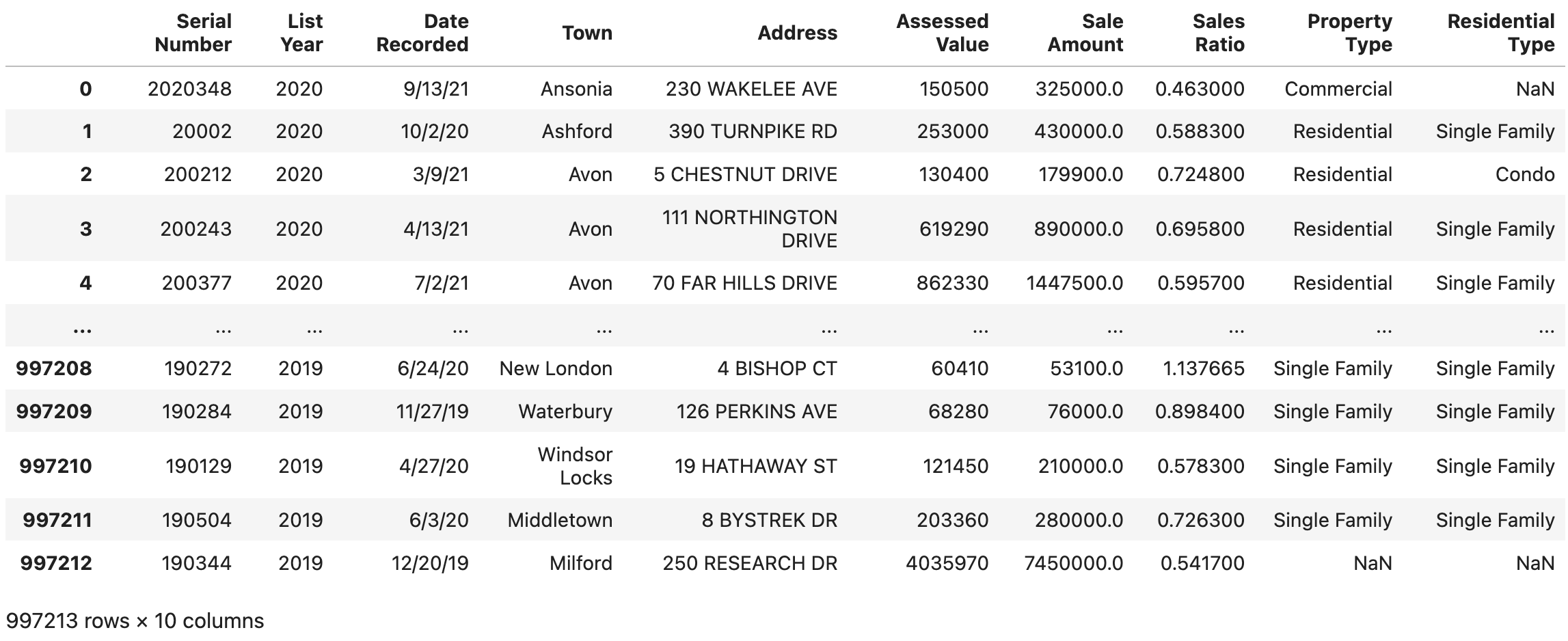}}
\caption{CT Real Estate Transaction 2001-2020}
\label{fig3}
\end{figure}

\begin{figure}[htbp]
\centerline{\includegraphics[width=\linewidth]{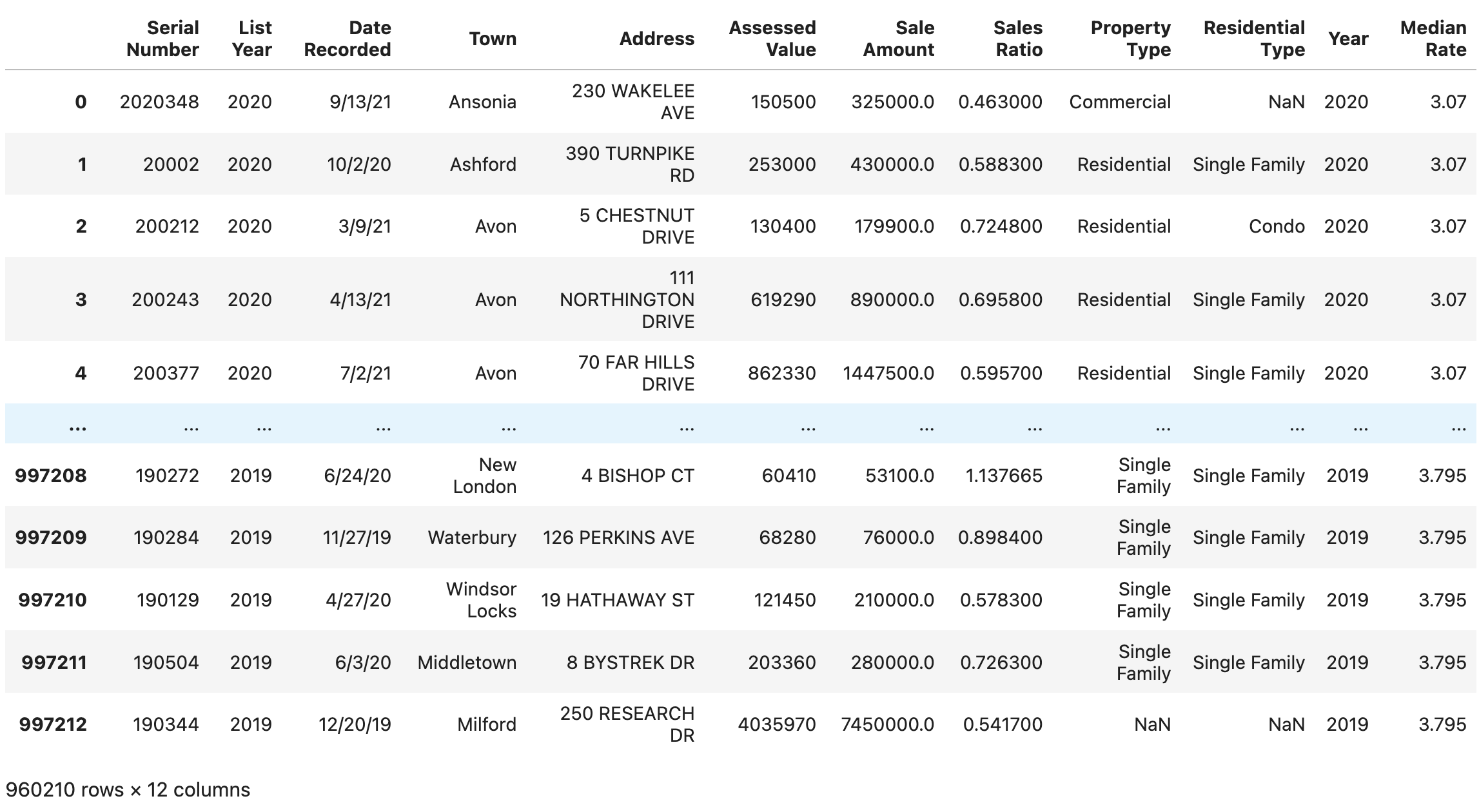}}
\caption{CT Real Estate With Interest Rate}
\label{fig4}
\end{figure}

2. Data Merging: The median interest rate per year was then merged with the real estate sales data (Figure~\ref{fig3}) based on the 'Year' column. This merge operation facilitated the integration of financial indicators with property-specific information, allowing for the subsequent analyses to consider both market conditions and individual property attributes. (Figure~\ref{fig4})

3. Town-Level Aggregation: For each town and each year within the real estate dataset, the median sales ratio and the number of sales were computed. This operation was designed to capture localized market trends and transaction volumes, which are indicative of the investment viability in different regions.

4. Investment Assessment: To simulate an investment scenario, a hypothetical investment amount of \$500,000 was considered. For each property, the assessed value that an investor could afford based on the sales ratio was calculated, along with the total payment over a 30-year mortgage period using the median interest rate for that year. This step involved applying the standard mortgage payment formula to the dataset and was instrumental in estimating the long-term financial commitment associated with each potential investment.

5. Buy Recommendation Labeling: The final preprocessing step involved the creation of a binary 'Buy Recommendation' label. A property was labeled with a '1' (recommend to buy) if the total payment minus the assessed value was less than the town's average for that metric; otherwise, it was labeled with a '0' (do not buy). This labeling was informed by the hypothesis that properties with total payments significantly exceeding the assessed value, relative to the local average, may not be prudent investments.

\begin{figure}[htbp]
\centerline{\includegraphics[width=\linewidth]{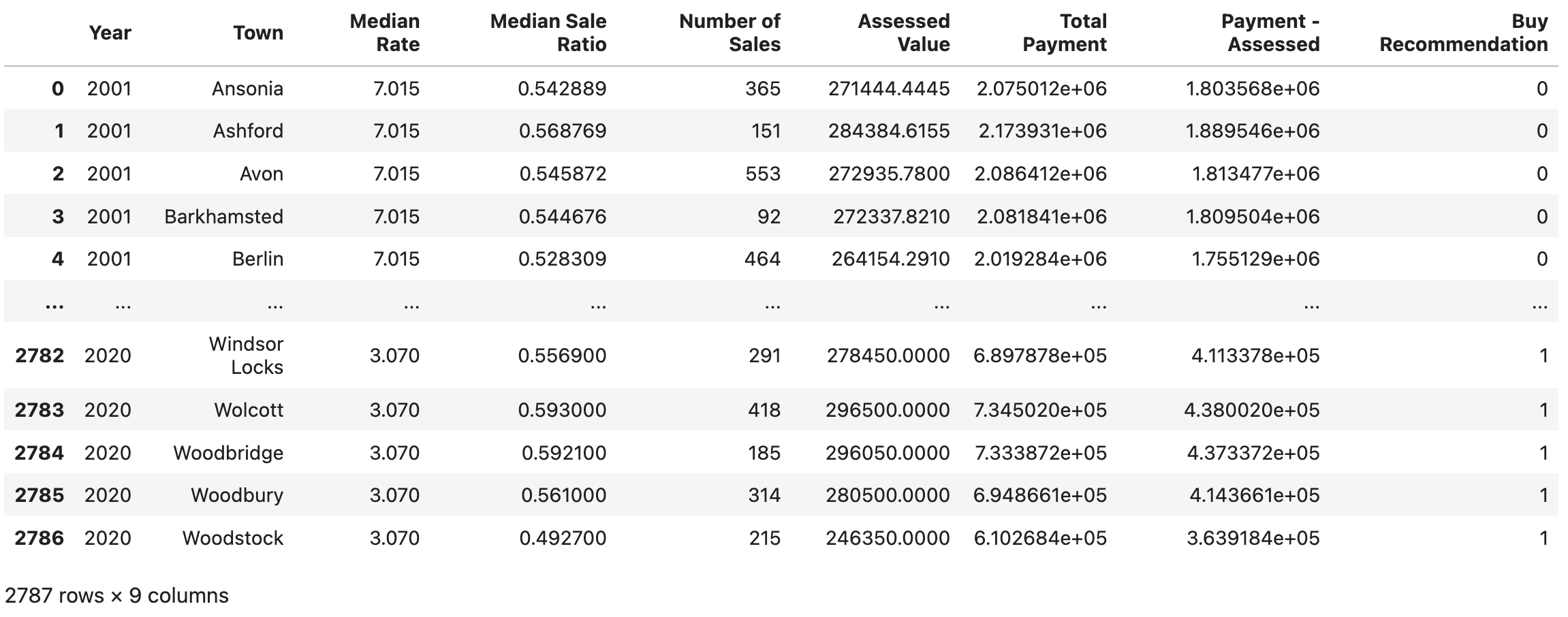}}
\caption{Preprocessed Data Set}
\label{fig}
\end{figure}

These preprocessing steps transformed the raw data into a structured form suitable for predictive analysis, enabling a robust comparison of the Linear Regression, Gaussian Elimination, and LU Decomposition algorithms in the subsequent stages of the study.

\section{Training and Testing}
The core of this research involved the application and evaluation of three distinct algorithms on the task of predicting buy recommendations for real estate investments. The dependent variable, 'Buy Recommendation', was a binary indicator of whether a real estate purchase should be advised (1) or not advised (0), based on a set of independent variables: 'Town', 'Year', 'Median Rate', and 'Median Sale Ratio'. The dataset was split into a training set to train the models and a testing set to evaluate their predictive performance.

\subsection{Linear Regression Model}

The Linear Regression model was implemented using the scikit-learn library, a choice made for its robustness and widespread use in the machine learning community. The model was trained on the specified features, and its performance was evaluated using standard metrics. The training score achieved was 0.7056, indicating a strong positive correlation between the model predictions and the actual values in the training set. The test score was 0.6359, signifying a good generalization of the model to unseen data. The Mean Squared Error (MSE) on the testing set was calculated to be 0.0902, demonstrating the model's accuracy in predictions. The computational efficiency of the Linear Regression model was noted, with a total runtime of 131 ms.

\begin{lstlisting}[language=Python, caption=Linear Regression Code]
one_hot_encoder = OneHotEncoder()
encoded_towns = one_hot_encoder.fit_transform(grouped['Town'].values.reshape(-1, 1)).toarray()
years = grouped['Year'].values.reshape(-1, 1)
interest_rates = grouped['Median\ Rate'].values.reshape(-1, 1)
Sale_Ratio = grouped['Median\ Sale\ Ratio'].values.reshape(-1, 1)
X = np.hstack([encoded_towns, years, interest_rates, Sale_Ratio])
y = grouped['Buy\ Recommendation'].values
X_train, X_test, y_train, y_test = train_test_split(X, y, test_size=0.25, random_state=42)
lin_reg = LinearRegression()
lin_reg.fit(X_train, y_train)
y_pred_test_lr = lin_reg.predict(X_test)
train_score_lr = r2_score(y_train, lin_reg.predict(X_train))
test_score_lr = r2_score(y_test, y_pred_test_lr)
mse_lr = mean_squared_error(y_test, y_pred_test_lr)
\end{lstlisting}

\subsection{Gaussian Elimination with Partial Pivoting}

Next, the Gaussian Elimination algorithm with partial pivoting \cite{b3} was applied to the same dataset. Despite its widespread use in numerical linear algebra for solving systems of linear equations, its application in predictive modeling presented challenges. The test score was considerably lower at 0.2510 compared to the Linear Regression model. The MSE increased to 0.1855, indicating a larger deviation of the predicted values from the actual values. The time complexity was also higher, with the Gaussian Elimination method taking 441 ms to run, which is more than thrice the time taken by the Linear Regression model.

\begin{lstlisting}[language=Python, caption=Gaussian Elimination Code]
if not isinstance(X_train, np.ndarray):
    X_train = X_train.toarray()
A = np.hstack([np.ones((X_train.shape[0], 1)), X_train, y_train.reshape(-1, 1)])
n = X_train.shape[1] + 1
reg_factor = 1e-10
for i in range(n):
    max_row = i + np.argmax(np.abs(A[i:, i]))
    pivot = A[max_row, i]
    if np.abs(pivot) < reg_factor:
        regularization = reg_factor if pivot >= 0 else -reg_factor
        A[max_row, i] += regularization
    else:
        regularization = 0
    A[[i, max_row]] = A[[max_row, i]]
    for k in range(i+1, A.shape[0]):
        factor = A[k, i] / (A[i, i] + regularization)
        A[k, i:] -= factor * A[i, i:]
x = np.zeros(n)
for i in range(n-1, -1, -1):
    if np.abs(A[i, i]) < reg_factor:
        raise ValueError(f'Pivot element too small for stable solution at row {i}.')
    x[i] = (A[i, -1] - np.dot(A[i, i+1:n], x[i+1:])) / A[i, i]
coefficients_ge = x
if not isinstance(X_test, np.ndarray):
    X_test = X_test.toarray()
X_test_augmented = np.hstack([np.ones((X_test.shape[0], 1)), X_test])
if not np.isnan(x).any() and not np.isnan(X_test_augmented).any():
    y_pred_test_ge = X_test_augmented.dot(x)
    residuals_test_ge = y_test - y_pred_test_ge
    r_squared = r2_score(y_test, y_pred_test_ge)
    mse = mean_squared_error(y_test, y_pred_test_ge)
\end{lstlisting}

\subsection{LU Decomposition Method}

The LU Decomposition method \cite{b3} was the third algorithm applied. This approach decomposes a matrix into lower and upper triangular matrices, which can then be used to solve systems of equations. The LU Decomposition method resulted in a test score of 0.6427, closely rivaling that of the Linear Regression model. Its MSE was 0.0885, slightly better than that of the Linear Regression model, suggesting a competitive accuracy in predictions. However, similar to Gaussian Elimination, the LU Decomposition method required 439 ms to execute, indicating a higher computational demand than Linear Regression.

\begin{lstlisting}[language=Python, caption=LU Decomposition Code]
def LUdecomp_without_pivot(a):
    n = len(a)
    L = np.eye(n) 
    U = a.copy()
    for k in range(n - 1):
        for i in range(k + 1, n):
            factor = U[i, k] / U[k, k]
            L[i, k] = factor  # Store the factor in L
            for j in range(k + 1, n):
                U[i, j] -= factor * U[k, j]
    return L, U
def forward_substitution(L, b):
    n = len(L)
    y = np.zeros(n)
    for i in range(n):
        y[i] = b[i] - np.dot(L[i, :i], y[:i])
    return y
def backward_substitution(U, y):
    n = len(U)
    x = np.zeros(n)
    for i in range(n-1, -1, -1):
        x[i] = (y[i] - np.dot(U[i, i+1:], x[i+1:])) / U[i, i]
    return x
def LU_solve(A, b):
    ATA = A.T @ A
    ATb = A.T @ b
    L, U = LUdecomp_without_pivot(ATA)
    y = forward_substitution(L, ATb)
    x = backward_substitution(U, y)
    return x
if not isinstance(X_train, np.ndarray):
    X_train = X_train.toarray()
A = np.hstack([np.ones((X_train.shape[0], 1)), X_train])
x = LU_solve(A, y_train)
coefficients_lu = x
A_test = np.hstack([np.ones((X_test.shape[0], 1)), X_test])
y_pred_test_lu = A_test.dot(x)
residuals_test_lu = y_test - y_pred_test_lu
test_score_lu = r2_score(y_test, y_pred_test_lu)
mse_lu = mean_squared_error(y_test, y_pred_test_lu)
\end{lstlisting}

\section*{Evaluation}
The evaluation of the three computational algorithms—Linear Regression (LR), Gaussian Elimination (GE) with partial pivoting, and LU Decomposition (LU)—was predicated on their underlying mathematical principles and their stability when handling numerical data. The inherent numerical instability in GE and LU, particularly when dealing with ill-conditioned matrices, could potentially affect the reliability of the results. In contrast, LR, as implemented in the scikit-learn library, incorporates numerical safeguards that enhance its stability and performance.

\subsection{Underlying Algorithms and Numerical Stability}

LR is a statistical method that estimates the coefficients of a linear equation, involving one or more independent variables that predict the outcome of a dependent variable. The scikit-learn implementation of LR employs techniques such as QR decomposition, which is less sensitive to numerical issues.

GE, a method for solving linear systems, is susceptible to numerical instability due to the potential for large round-off errors if the pivot elements are very small. Partial pivoting mitigates this to some extent by swapping rows to use the largest absolute pivot element, but it does not guarantee stability in all cases.

LU Decomposition, a matrix factorization technique, decomposes a matrix into a lower triangular matrix (L) and an upper triangular matrix (U). This method can be more stable than GE due to its systematic approach to decomposition, which can reduce the impact of round-off errors.

\subsection{Performance Discrepancy Analysis}

The observed performance discrepancy between the three algorithms can be attributed to several factors. LR's implementation in scikit-learn is highly optimized for predictive modeling, making it robust against common numerical issues. GE's performance was less satisfactory, likely due to cumulative round-off errors and potential instability when the system's matrix is not well-conditioned.

\subsection{Residual Analysis}
\begin{figure}[htbp]
\centerline{\includegraphics[width=\linewidth]{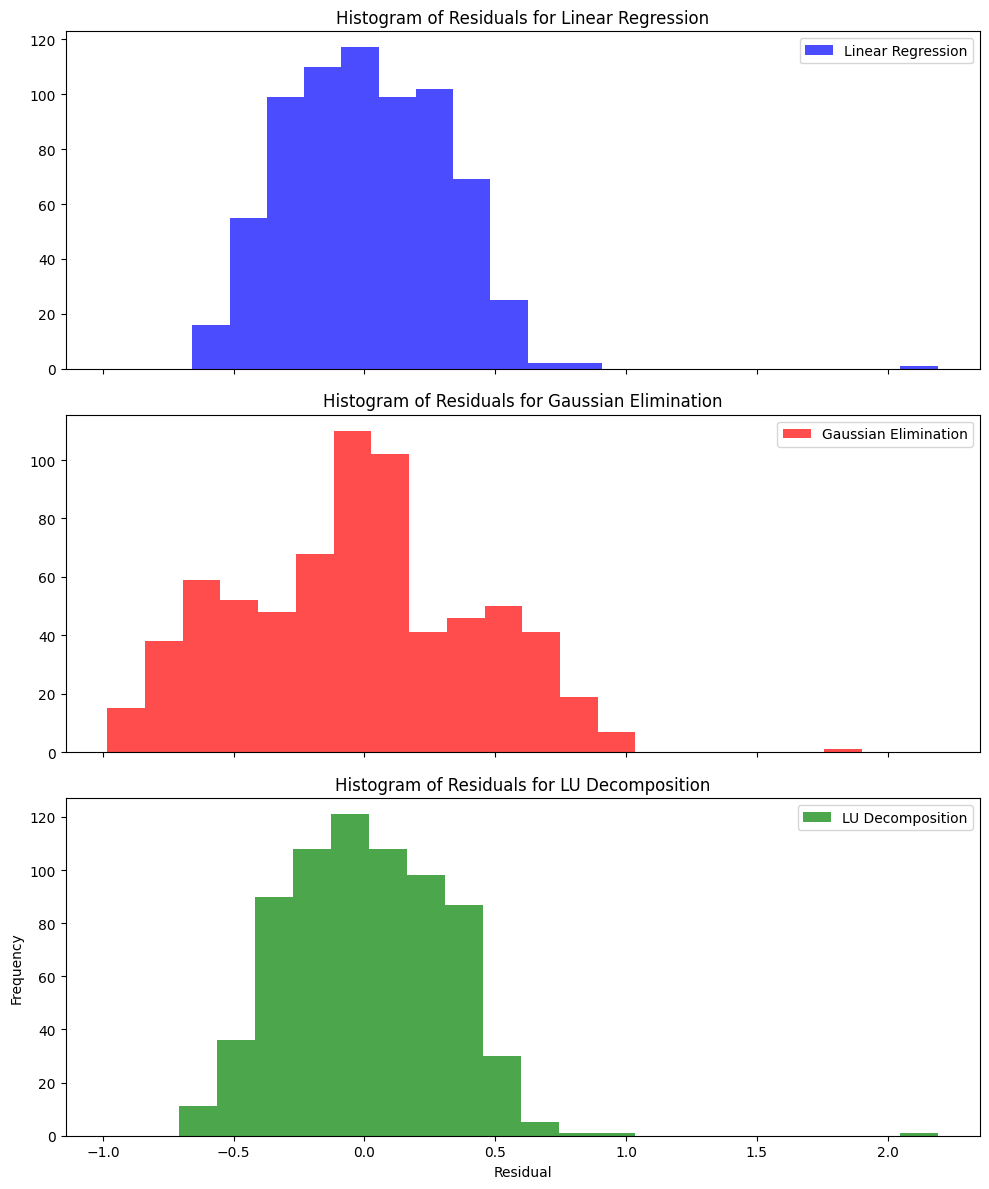}}
\caption{Residuals Comparison}
\label{fig6}
\end{figure}

The histogram of residuals for each algorithm (Figure~\ref{fig6}) offers insight into the distribution of prediction errors. The residuals from the Linear Regression model demonstrate a relatively symmetrical distribution, indicating that the model is consistent in its predictive performance across the dataset. Gaussian Elimination, however, shows a skewed residual distribution, suggesting a systematic bias in the predictions or a potential misfit. The LU Decomposition model's residuals are more symmetric, similar to Linear Regression, but with a slight skew. This skewness in Gaussian Elimination and LU Decomposition could be due to the numerical instability inherent in these methods when applied to datasets with high condition numbers.

\subsection{Coefficient Analysis}
\begin{figure}[htbp]
\centerline{\includegraphics[width=\linewidth]{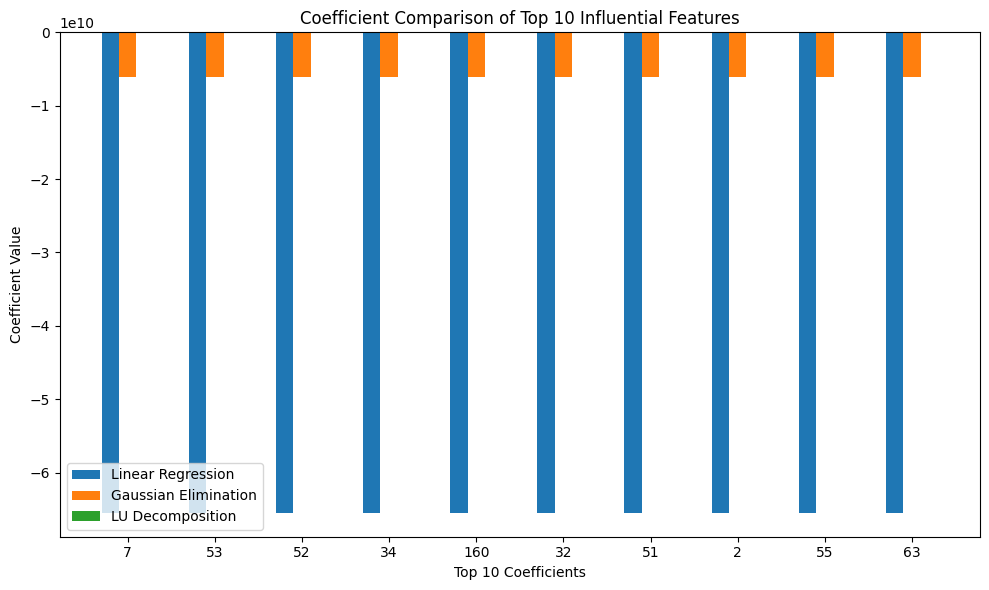}}
\caption{Top 10 Coefficient Comparison}
\label{fig7}
\end{figure}

The coefficient comparison (Figure~\ref{fig7}) highlights the stark differences in the magnitude of coefficients estimated by each model. The Linear Regression model maintains coefficient magnitudes within a narrower range, which could indicate a more stable estimation process. In contrast, the coefficients derived from Gaussian Elimination and LU Decomposition are significantly larger, pointing to potential overfitting and numerical instability issues, especially when considering the extremely high condition number of 3.14×10**17 of the matrix involved. This condition number, which is a measure of the sensitivity of the function's output to its input, suggests that the dataset is ill-conditioned for these linear models, resulting in unreliable and unstable coefficient estimates.

\subsection{Superior Performance of LU Decomposition}

LU Decomposition's superior performance compared to GE could be due to the structured way in which LU handles matrix decomposition, which can be less prone to amplifying errors in the presence of small pivot elements. Moreover, LU's two-step solution process—forward elimination followed by backward substitution—may offer better control over numerical precision.

\subsection{Runtime Discrepancy Analysis}

The disparity in runtime between Linear Regression (131 ms), Gaussian Elimination (441 ms), and LU Decomposition (439 ms) is primarily due to the inherent computational complexity and the efficiency of the underlying numerical procedures. Linear Regression, optimized within scikit-learn, benefits from advanced matrix operations that are highly optimized for speed, especially on large datasets. This optimization often includes the use of vectorized operations and linear algebra routines that can process data in bulk rather than iteratively.

On the other hand, Gaussian Elimination and LU Decomposition are more computationally intensive by nature. The need for row manipulations, particularly with partial pivoting in Gaussian Elimination, introduces significant computational overhead. Although LU Decomposition theoretically simplifies the process by breaking it down into two triangular matrix solutions, the lack of optimization in the matrix decomposition and solving phases can result in runtimes that closely mirror those of Gaussian Elimination. This evaluation highlights the crucial role of algorithmic efficiency in data processing and the potential benefits of leveraging specialized, optimized algorithms for predictive modeling in real estate and other data-intensive domains.

\subsection{Impact of Real Estate Dataset Complexity}

The performance of all three algorithms was not optimal, suggesting that the complexity of the real estate dataset may exceed the modeling capacity of linear methods. Real estate markets are influenced by a multitude of factors, many of which are nonlinear and interdependent. The assumption of linearity in the relationship between the features and the target variable may be too simplistic for this domain. Additionally, the presence of outliers, multicollinearity, and heteroscedasticity could further degrade the model's performance.

\section*{Conclusion}
This study compared the efficacy of Linear Regression, Gaussian Elimination with partial pivoting, and LU Decomposition in predicting real estate investment decisions, revealing Linear Regression's superiority in both performance and computational efficiency. The investigation highlighted the challenges linear models face in capturing the complex dynamics of the real estate market, suggesting a potential mismatch between model simplicity and market complexity. Future work should focus on deploying more sophisticated, possibly nonlinear, machine learning models that can better handle the multifaceted nature of real estate data, with an emphasis on exploring features beyond the scope of traditional valuation metrics. The insights garnered point toward an evolving landscape in predictive real estate analytics, where advanced computational methods could significantly enhance investment decision-making processes.

\end{document}